\documentclass{article}

\usepackage[dblblindworkshop, final]{neurips_2025}
\workshoptitle{DiffCoALG: Differentiable Learning of Combinatorial Algorithms}

\usepackage[utf8]{inputenc} 
\usepackage[T1]{fontenc}    
\usepackage{hyperref}       
\usepackage{url}            
\usepackage{booktabs}       
\usepackage{amsfonts}       
\usepackage{nicefrac}       
\usepackage{microtype}      
\usepackage{xcolor}         

\usepackage{amsmath}
\usepackage{graphicx}
\usepackage{verbatim}
\usepackage{float}
\usepackage{multirow}

\title{Offline Decision Transformers for Neural Combinatorial Optimization: Surpassing Heuristics on the Traveling Salesman Problem}

\author{%
  Hironori Ohigashi \\
  Panasonic Connect Co., Ltd. \\
  Japan \\
  \texttt{ohigashi.hironori@jp.panasonic.com} \\
  \And
  Shinichiro Hamada \\
  Panasonic Connect Co., Ltd. \\
  Japan \\
  \texttt{hamada.shinichiro001@jp.panasonic.com} \\
}

\begin{document}

\maketitle

\begin{abstract}
Combinatorial optimization problems like the Traveling Salesman Problem are critical in industry yet NP-hard. Neural Combinatorial Optimization has shown promise, but its reliance on online reinforcement learning (RL) hampers deployment and underutilizes decades of algorithmic knowledge. We address these limitations by applying the offline RL framework, Decision Transformer, to learn superior strategies directly from datasets of heuristic solutions—aiming not only to imitate but to synthesize and outperform them. Concretely, we (i) integrate a Pointer Network to handle the instance-dependent, variable action space of node selection, and (ii) employ expectile regression for optimistic conditioning of Return-to-Go, which is crucial for instances with widely varying optimal values. Experiments show that our method consistently produces higher-quality tours than the four classical heuristics it is trained on, demonstrating the potential of offline RL to unlock and exceed the performance embedded in existing domain knowledge. 
\footnote{Our code is available at \url{https://github.com/PanasonicConnect/dt-tsp}.}
\end{abstract}

\section{Introduction} \label{sec:introduction}

Combinatorial optimization problems (COP) have garnered significant attention in various industries, including logistics, manufacturing, and communication network design. Many of these problems are NP-hard, making it extremely difficult to find exact optimal solutions efficiently \cite{nphard}. Consequently, heuristics and metaheuristics have been studied for many years as practical methods for finding approximate solutions \cite{approx, metaheu}. However, these methods suffer from challenges in generalizability and scalability, as computational costs increase with problem size and they often require parameter tuning \cite{metaheu-challenge}.

In recent years, advancements in deep learning have given rise to Neural Combinatorial Optimization (NCO) \cite{bello,kool,learning-tsp,pomo,h-tsp,jieyi,unico}. However, a predominant approach in NCO relies on reinforcement learning (RL), which requires collecting data through interaction with an environment. This online learning process presents practical challenges for real-world deployment, as it requires either resource-intensive data acquisition from real environments, or designing a surrogate virtual environment, a task complicated by the implicit knowledge involved \cite{online-challenge}. Moreover, leveraging the rich knowledge from domain-specific heuristics and human experts remains a significant, yet often unaddressed, challenge for these methods.

To address these challenges, we propose applying the Decision Transformer (DT) \cite{dt}, an offline RL framework proven in other domains \cite{mgdt,dt4robot,dt4trading,dt4planing,dt4driving}, to learn from pre-existing datasets of heuristic solutions. This approach enables the use of algorithmic and expert domain knowledge as valuable data for a neural network to learn solution methods.
In this paper, we propose a novel formulation for applying the DT to the Traveling Salesman Problem (TSP). As the standard DT is not designed for node-selection tasks whose action spaces lack semantic consistency, we integrate the Pointer Network \cite{pointernet} into the action selection mechanism for TSP. We further equip the DT with a mechanism, inspired by \cite{mgdt, edt}, to predict the highest possible returns for each instance, in order to address COP where the optimal reward varies significantly across instances.

Our contributions are: 1) a novel DT framework for TSP that consistently generates solutions superior to the heuristic data it was trained on; 2) a clear demonstration that conditioning the model with appropriate Return-to-Go (RTG) is critical for outperforming behavior cloning; and 3) validation that optimistic RTG prediction, via expectile regression, enhances solution quality.

These results suggest that offline RL frameworks like the DT can be a powerful tool for utilizing existing domain knowledge to generate innovative solutions for complex COP.

\section{Methods} \label{sec:methods}

\subsection{TSP formulation}

This paper focuses on the 2D Euclidean TSP. The problem is defined on an undirected graph $G=(V, E)$ consisting of a set of nodes $V=\{v_i\}_{i=1}^N$ and a set of edges $E = \{(v_i, v_j) | i<j, 1 \le i, j \le N\}$. Here, $N$ is the total number of nodes, and the travel cost $cost(v_i, v_j)$ for each edge is given by the Euclidean distance between the nodes. A salesman starts from a special depot node $v_d$, visits every node exactly once, and returns to the start, forming a Hamiltonian cycle. The objective is to minimize the total cost of this tour, denoted as $L(\boldsymbol{\sigma})$, where $\boldsymbol{\sigma}$ is the tour route. The total cost is expressed by the following equation:

\begin{equation}
L(\boldsymbol{\sigma}) = \sum_{i=1}^{N-1} cost(\sigma_i, \sigma_{i+1}) + cost(\sigma_N, \sigma_1)
\end{equation}

Here, $\sigma_i$ is the $i$-th node in the tour, and $\sigma_1=v_d$.

\subsection{Application of DT}

Many constructive NCO studies \cite{bello,kool,learning-tsp,pomo,h-tsp,jieyi,unico} formulate TSP as a Markov Decision Process (MDP) where a node to visit next is selected at each time step. In this approach, the state at time $t$ for a TSP instance graph $G$ is defined as a partial tour $\boldsymbol{\sigma}_{1:t} = (\sigma_1, \sigma_2, ..., \sigma_t)$ consisting of visited nodes. The action is the selection of the next node $\sigma_{t+1}$, and this decision is made by a deep learning model with parameters $\theta$, $\pi_\theta(\sigma_{t+1}|\sigma_{1:t},G)$. The model is trained using a RL framework with a reward equal to the negative of the total tour cost, $-L(\boldsymbol{\sigma})$.

While many NCO methods formulate TSP as a MDP, we adopt the DT's sequence modeling approach. We model trajectories $\tau = (\ldots, o_t, \hat{R}_t, a_t, \ldots)$, where $\hat{R}_t$ is the RTG, $o_t$ is the observation, and $a_t$ is the action.

To adapt this framework to TSP, we redefine $o_t, \hat{R}_t, a_t$ as follows: $o_t$ is the embedding vector $\boldsymbol{f}^{e}_{t}$ computed by the model's Encoder, corresponding to the node $\sigma_t$ visited at time $t$. This Encoder, following the architecture of Kool et al. \cite{kool}, uses a transformer Encoder with node coordinate information as input to compute node embedding vectors. $\hat{R}_t$ is the negative of the total cost of the completed tour $\boldsymbol{\sigma}$ at the final time step $T$, i.e., $-L(\boldsymbol{\sigma})$. $a_t$ represents the index of the next node $\sigma_{t+1}$ to be visited.

\begin{figure}
  \centering
  \includegraphics[width=0.9\textwidth]{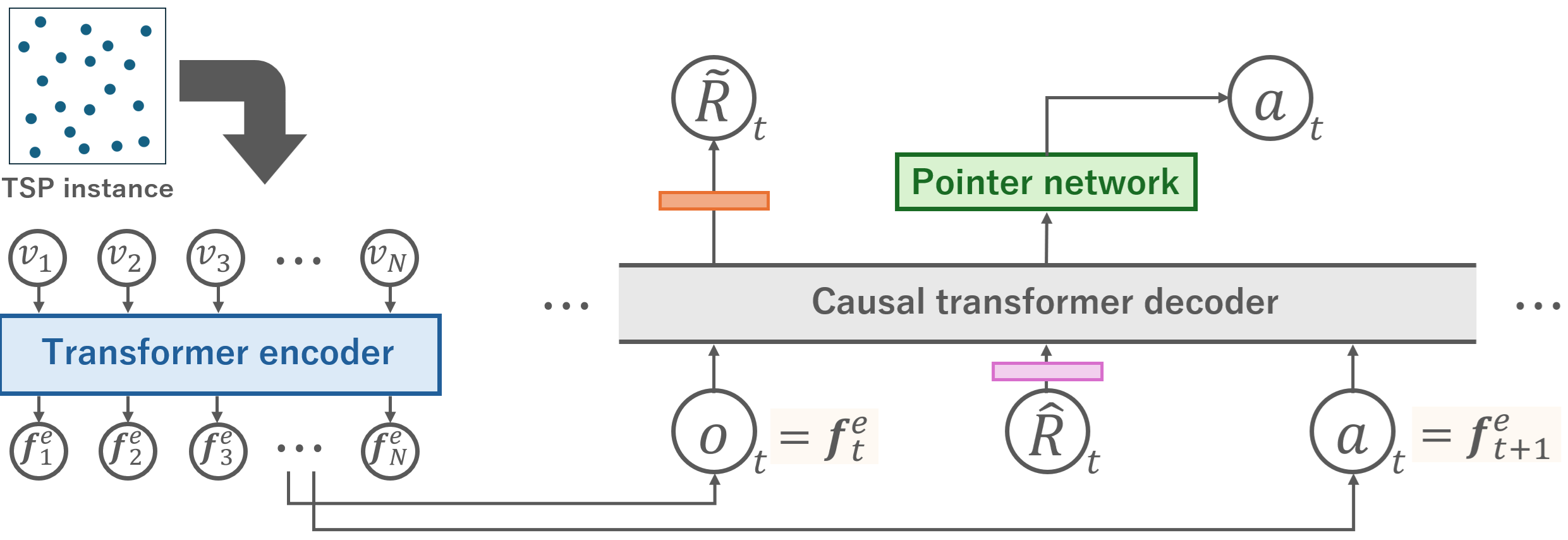}
  \caption{Overview of the proposed method's architecture. The transformer encoder calculates a node embedding vector from the coordinates of the graph nodes. As observation $o_t$ and action $a_t$, the embedding vectors $\boldsymbol{f}^{e}_{t}$ and $\boldsymbol{f}^{e}_{t+1}$ of the nodes transitioned at time $t$ and $t+1$, respectively, are input to the causal transformer decoder along with the RTG $\hat{R}_t$. The value $\tilde{R}_t$ is the RTG prediction by the causal transformer decoder.}
  \label{fig:1}
\end{figure}

The overall architecture of the proposed method is shown in Figure \ref{fig:1}.

\subsection{Action representation via Pointer Network}

Standard DT actions assume semantic consistency (e.g., "up" or "down"), which does not hold for node indices in TSP as their spatial meaning varies per instance. 
To address this, as shown in Figure \ref{fig:1}, we introduce a Pointer Network \cite{pointernet} to the output of the causal transformer decoder \cite{transformer}. A Pointer Network generates an output sequence by "pointing" to elements within the input sequence. This approach modifies the DT's action head to output pointers to the graph nodes of the TSP instance, rather than probabilities over a fixed set of actions. In this paper, the pointer is calculated by an attention mechanism as follows:

\begin{align}
    u_i^{t+1}&=\frac{\boldsymbol{h}_t^{a}\cdot\boldsymbol{f}_i^{e}}{\sqrt{d}}\\
    p(v_i|\boldsymbol{\sigma}_{1:t},G)&=
    \begin{cases}
        0&\text{if }v_i\in\boldsymbol{\sigma}_{1:t}\\
        \text{softmax}(u_i^{t+1})&\text{otherwise}
    \end{cases}
\end{align}

Here, $\boldsymbol{f}_i^{e}$ is the node embedding vector for node $v_i$ computed by the Encoder, $\boldsymbol{h}_t^{a}$ is the hidden state of the action head at time $t$, and $d$ is the dimension of $\boldsymbol{h}_t^{a}$.
  
Similarly, when providing the action $a_t$ as input to the DT, we convert the selected node $\sigma_{t+1}$ to its node embedding vector $\boldsymbol{f}_{t+1}^{e}$ instead of using its index.

\subsection{RTG prediction}

Methods using the transformer architecture are often applied to tasks where rewards are relatively predictable, such as games and robot control. However, in NP-hard COP like TSP, the optimal reward varies greatly for each instance. If a uniform RTG is set, the model might treat it as an "extrapolated input" outside its learning range. For example, providing a target value of 2.5 for an instance with an optimal value of 3.8 could lead to performance degradation.

To solve this issue, we introduce a mechanism for dynamically predicting the RTG, inspired by frameworks like Multi-game DT \cite{mgdt} and Elastic DT \cite{edt}. In this approach, the RTG is predicted from the graph information of the TSP instance, and this value is then used for predicting the solution. Specifically, the predicted RTG value $\tilde{R}_t$ is output by the DT model as $\pi_{\theta}(\tilde{R}_{t+1} | \tau_{1:t}, G)$ with the graph information $G$ and the partial sequence $\tau_{1:t}$ as inputs. This predicted value is then used as the RTG for the next action prediction.

The predicted RTG $\tilde{R}_t$ should ideally reflect the maximum achievable return. For this reason, we use expectile regression \cite{expectile1,expectile2} to train the predicted value. The loss function for this is defined by the following equation:

\begin{equation}
L_{\alpha}^2 (\hat{R}_t, \tilde{R}_t) = | \alpha - \mathbf{1}(\hat{R}_t < \tilde{R}_t)| \cdot (\hat{R}_t - \tilde{R}_t)^2
\end{equation}

Here, $\alpha$ is a hyperparameter that controls the weighting of the error. Specifically, if $\alpha > 0.5$, the model places more emphasis on under-prediction errors, while if $\alpha < 0.5$, it emphasizes over-prediction errors. When $\alpha = 0.5$, it becomes equivalent to the squared error.

\subsection{Learning objective and loss function}

Our model is trained via multi-task learning, minimizing a combined loss $L_{total} = L_{CE} + c \cdot L_{\alpha}^2$, where $L_{CE}$ is the cross-entropy loss for the action prediction (node selection), and $L_{\alpha}^2$ is the expectile regression loss for predicting the RTG.  The hyperparameter $\alpha$ encourages optimistic RTG prediction, while $c$ balances the two tasks.

\section{Experimental results}

\subsection{Experimental setup} \label{sec:experimental_setup}

In this study, we focus on 2D euclidean TSP to validate the effectiveness of the proposed method.

For our dataset, we used the 2D Euclidean TSP instances with $N=20, 50, 100$, which are available from Joshi et al. \cite{learning-tsp}. In these instances, the coordinates of each node are sampled from a uniform distribution over the unit square $[0, 1]^2$. For each node, we prepared 1,000,000 instances for training, 10,000 for validation, and 10,000 for testing. Since the validation datasets for $N=50, 100$ were not provided, we generated them anew. Following the methodology of Joshi et al. \cite{learning-tsp}, the solution data was generated using four heuristics: Nearest Neighbor (NN), Nearest Insertion (NI), Farthest Insertion (FI) \cite{nn-ni-fi}, and Simulated Annealing (SA) \cite{sa}.

For our DT model, we used a 2-layer transformer encoder and a 2-layer causal transformer decoder, inspired by the Elastic DT. Each layer was set with a hidden dimension $d_{model}=128$ and 8 heads.
We used the Schedule-Free AdamW optimizer \cite{schedulefree} with a learning rate of 0.0025 and a batch size of 1000.
For the total loss function $L_{total} = L_{CE} + c \cdot L_{\alpha}^2$, we set the hyperparameter $c$ to 0.5 and $\alpha$ to 0.99. The model was trained for 2000 epochs, and we selected the model from the epoch where the validation loss was minimal. Further details on the experimental setup are provided in Appendix \ref{sec:appendix_experimental_details}.

To evaluate the performance of our heuristics and model, we calculated the optimality gap (\%) against the exact optimal solutions provided by Joshi et al. \cite{learning-tsp} on the test set of 10,000 instances. The optimality gap is defined as follows:

\begin{equation}\label{eq:optimal_gap}
\text{optimality gap} (\%) = \frac{1}{M} \sum_{m=1}^{M} \frac{L(\boldsymbol{\sigma}^m_{\mathrm{pred}})-L(\boldsymbol{\sigma}^m_{\mathrm{opt}})}{L(\boldsymbol{\sigma}^m_{\mathrm{opt}})} \times 100
\end{equation}

where $M$ denotes the number of test instances and, for instance $m$, $L(\boldsymbol{\sigma}^m_{\mathrm{opt}})$ and $L(\boldsymbol{\sigma}^m_{\mathrm{pred}})$ denote the costs of the optimal solutions and model-predicted solutions, respectively.

\subsection{Prediction performance of the proposed method}

Following the experimental setup described in Section \ref{sec:experimental_setup}, we trained our proposed method on each training dataset generated by the NN, NI, FI, and SA heuristics. We then evaluated the performance of each resulting model by calculating the optimality gap of its predicted solutions. This entire procedure was conducted for problem sizes of $N=20, 50, 100$.

Table \ref{tab:main} shows that our method consistently outperformed the solutions from all training heuristics. In particular, the most notable improvement, approximately a twofold increase over the original heuristic, was observed for the dataset generated from SA. We hypothesize that this is because the stochastic nature of SA produces highly diverse solution patterns, which enabled our DT model to better stitch together sub-optimal segments from different solution trajectories. The limited improvement on the NN dataset will be discussed in detail in the following section.

\begin{table}
\centering
\caption{Optimality gap (\%) and its standard deviation for each method on the test dataset. The "Data" column indicates the heuristic method used for training. The "Method" column represents the original heuristic method (Original), behavior cloning (BC), and our proposed method (DT).}
\label{tab:main}
\begin{tabular}{ll|rrr}
\toprule
\multicolumn{1}{l}{Data} & \multicolumn{1}{l}{Method} & \multicolumn{1}{c}{$N=20$} & \multicolumn{1}{c}{$N=50$} & \multicolumn{1}{c}{$N=100$} \\
\midrule
\multirow{3}{*}{NN} & Original & 17.24 $\pm$ 10.24 & 22.73 $\pm$ 8.21 & 24.81 $\pm$ 6.47\\
\cmidrule(lr){2-5}
 & BC & 17.28 $\pm$ 10.23 & 22.74 $\pm$ 8.21 & \textbf{24.75} $\pm$ 6.50\\
 & DT (Ours) & \textbf{16.73} $\pm$ 10.07 & \textbf{22.60} $\pm$ 8.17 & 24.76 $\pm$ 6.48\\
\midrule
\multirow{3}{*}{NI} & Original & 13.24 $\pm$ 6.99 & 19.12 $\pm$ 4.78 & 21.77 $\pm$ 3.43 \\
\cmidrule(lr){2-5}
 & BC & 12.26 $\pm$ 7.13 & 18.20 $\pm$ 5.44 & 22.00 $\pm$ 5.29 \\
 & DT (Ours) & \textbf{6.43} $\pm$ 4.86 & \textbf{14.98} $\pm$ 4.91 & \textbf{19.84} $\pm$ 4.85 \\
\midrule
\multirow{3}{*}{FI} & Original & 2.36 $\pm$ 2.91 & 5.62 $\pm$ 3.12 & 7.62 $\pm$ 2.55 \\
\cmidrule(lr){2-5}
 & BC & 1.85 $\pm$ 2.55 & 3.70 $\pm$ 2.54 & 5.22 $\pm$ 2.22 \\
 & DT (Ours) & \textbf{1.30} $\pm$ 2.27 & \textbf{3.14} $\pm$ 2.30 & \textbf{4.73} $\pm$ 2.12 \\
\midrule
\multirow{3}{*}{SA} & Original & 1.51 $\pm$ 2.79 & 4.41 $\pm$ 3.16 & 12.36 $\pm$ 3.66 \\
\cmidrule(lr){2-5}
 & BC & 0.98 $\pm$ 1.87 & 2.93 $\pm$ 2.57 & 10.31 $\pm$ 4.74 \\
 & DT (Ours) & \textbf{0.83} $\pm$ 1.60 & \textbf{2.39} $\pm$ 2.10 & \textbf{6.07} $\pm$ 3.11 \\
\bottomrule
\end{tabular}
\end{table}

\subsection{Performance comparison with behavior cloning}

To investigate the importance of appropriate RTG conditioning, we compared our method with behavior cloning. In behavior cloning, we used the model of the proposed method and set the RTG to 0 during both training and inference. 

Table \ref{tab:main} shows that our RTG-conditioned method generally outperformed behavior cloning. This confirms that RTG is essential for enabling the exploration required to find superior solutions, rather than merely replicating actions from the training data. The slight performance degradation of our proposed method compared to behavior cloning in the NN case for $N=100$ is likely due to the characteristics of NN. NN is a simple greedy algorithm that always selects the nearest neighbor node at each step, resulting in a lack of diversity in its action patterns. Consequently, the model could only learn a single action pattern from NN and had little opportunity to learn exploratory paths to better solutions. Therefore, no significant improvement over behavior cloning was observed. Additionally, since the proposed method involves the additional task of RTG prediction, this additional complexity may have prevented it from surpassing the performance of behavior cloning, which faithfully reproduces the simple action pattern.

\subsection{Effect of expectile regression on RTG}

To evaluate the effect of using expectile regression for RTG prediction, we conducted an ablation study. We compared our approach against baselines using fixed RTG targets: 0, used as a sufficiently high constant (following \cite{dt}), and the average RTG from the training data. Additionally, we analyzed the impact of the hyperparameter $\alpha$ from the RTG loss function, testing $\alpha=0.7, 0.99$ in addition to $\alpha=0.5$, which corresponds to standard regression using a squared error loss. Due to computational constraints, these experiments were performed exclusively on the datasets for $N=20, 50$.

Table \ref{tab:exp} shows that predicting the RTG generally yields superior solutions compared to using fixed-value targets. More importantly, the results clearly show that using expectile regression with $\alpha > 0.5$ consistently outperforms the squared error case ($\alpha=0.5$). Performance also tends to improve as $\alpha$ increases.
This improvement can be attributed to the mechanism of expectile regression: for $\alpha > 0.5$, under-prediction errors are weighted more heavily, which encourages the model to learn to predict higher RTGs. This optimistic prediction, in turn, allows the model to predict achievable, better solutions within the policies learned from the training data. This finding emphasizes the importance of focusing on the "best trajectories" within the data, rather than merely imitating the data distribution, when learning from offline dataset to surpass existing solutions.

\begin{table}
    \centering
    \caption{Optimality gap (\%) for different RTG targets (Fixed vs. Predicted) and values of $\alpha$ on $N=20, 50$.}
    \label{tab:exp}
    \begin{tabular}{ll|rr|rrr}
        \toprule
        & & \multicolumn{2}{c}{Fixed RTG} & \multicolumn{3}{c}{Predicted RTG using expectile regression} \\
        \multicolumn{1}{l}{$N$} & \multicolumn{1}{l}{Data} & \multicolumn{1}{c}{0} & \multicolumn{1}{c}{mean of data} & \multicolumn{1}{c}{$\alpha=0.50$} & \multicolumn{1}{c}{$\alpha=0.70$} & \multicolumn{1}{c}{$\alpha=0.99$} \\
        \midrule
        20 & NN & 67.90 $\pm$ 29.58 & 17.21 $\pm$ 10.17 & 17.22 $\pm$ 10.17 & 17.14 $\pm$ 10.13 & \textbf{16.73} $\pm$ 10.07 \\
         & NI & 68.39 $\pm$ 26.26 & 13.12 $\pm$ 7.62 & 13.09 $\pm$ 6.29 & 11.02 $\pm$ 5.96 & \textbf{6.43} $\pm$ 4.86 \\
         & FI & 89.69 $\pm$ 31.21 & 3.33 $\pm$ 4.78 & 1.70 $\pm$ 2.33 & 1.61 $\pm$ 2.27 & \textbf{1.30} $\pm$ 2.27 \\
         & SA & 106.03 $\pm$ 31.37 & 2.52 $\pm$ 4.50 & 0.97 $\pm$ 1.77 & 0.84 $\pm$ 1.50 & \textbf{0.83} $\pm$ 1.60 \\
        \midrule
        50 & NN & 213.67 $\pm$ 41.84 & 22.71 $\pm$ 8.18 & 22.69 $\pm$ 8.17 & 22.66 $\pm$ 8.17 & \textbf{22.60} $\pm$ 8.17\\
         & NI & 159.85 $\pm$ 41.00 & 18.15 $\pm$ 5.27 & 18.04 $\pm$ 5.34 & 17.69 $\pm$ 5.24 & \textbf{14.98} $\pm$ 4.91 \\
         & FI & 102.97 $\pm$ 34.79 & 4.16 $\pm$ 2.78 & 3.93 $\pm$ 2.48 & 3.75 $\pm$ 2.42 & \textbf{3.14} $\pm$ 2.30 \\
         & SA & 258.96 $\pm$ 45.99 & 3.45 $\pm$ 2.99 & 3.17 $\pm$ 2.55 & 2.90 $\pm$ 2.39 & \textbf{2.39} $\pm$ 2.10 \\
        \bottomrule
    \end{tabular}
\end{table}

\subsection{Exploring optimal RTG}

We investigated whether the RTG predictions of our model were sufficiently optimistic or if they could be improved. To this end, we conducted an experiment where a constant offset was systematically added to the RTG predictions during inference on the test data. This experiment aimed to determine if artificially inflating the RTG targets could compensate for potential underestimation by the model and thus lead to superior performance.

Table \ref{tab:offset} shows the optimality gaps relative to the original heuristic solutions, achieved by applying the optimal offset (value in parentheses) to the model trained on each heuristic.
In many cases, adding an offset resulted in better solutions than the original proposed method. This suggests that the RTGs predicted by our model may still be conservative, underestimating the target values required to elicit the best possible solutions. This trend became more pronounced with increasing problem size, indicating a greater difficulty in accurately predicting the optimal returns for large-scale instances.

\begin{table}
    \centering
    \caption{Optimality gap (\%) with optimal RTG offsets. The applied offset values are shown in parentheses.}
    \label{tab:offset}
    \begin{tabular}{l|rrr}
        \toprule
        Data & \multicolumn{1}{c}{$N=20$ (offset)} & \multicolumn{1}{c}{$N=50$ (offset)} & \multicolumn{1}{c}{$N=100$ (offset)}\\
        \midrule
        NN & 15.82 $\pm$ 9.96 (1.00) & 22.45 $\pm$ 8.14 (1.00) & 24.58 $\pm$ 6.49 (2.00)\\
        NI &  3.94 $\pm$ 4.34 (0.50) & 10.40 $\pm$ 5.10 (2.00) & 16.50 $\pm$ 5.06 (2.00)\\
        FI &  1.29 $\pm$ 2.21 (-0.05) &  3.07 $\pm$ 2.37 (0.20) &  4.51 $\pm$ 2.20 (0.50)\\
        SA &  0.79 $\pm$ 1.49 (-0.10) &  2.39 $\pm$ 2.10 (0.00) &  5.32 $\pm$ 2.87 (0.50)\\
        \bottomrule
    \end{tabular}
\end{table}

\section{Discussion and Conclusion} \label{sec:discussion_and_conclusion}

In this study, we proposed a new approach to solve the TSP by applying the offline RL framework DT to learn from existing heuristic solutions. Our experimental results demonstrated that the proposed method consistently outperform the solutions generated by the NN, NI, FI, and SA heuristics used for training. This result validates the fundamental concept of our approach: leveraging existing algorithmic knowledge as data to acquire a policy that surpasses it.

At the core of our method's success is goal-conditioned learning via RTG, which, unlike behavior cloning, learns the relationship between actions and outcomes. This enables the model to explore and generate novel, higher-quality solutions. Furthermore, our results with expectile regression show that setting optimistic goals—aiming for performance beyond the training data's average—is a key driver for this improvement, emphasizing the importance of focusing on the "best trajectories" within the offline dataset.

Our study also highlights several limitations and avenues for future work. The observation that adding a manual offset to the RTG improved performance suggests our prediction mechanism can be refined, especially for larger instances. Performance also depends on the training data's quality and diversity, as seen with the simple NN heuristic. Future work could explore training on more diverse datasets combining multiple heuristics or expert human solutions to learn a richer policy and tackle implicit real-world knowledge.

In conclusion, this study demonstrates that offline learning with the DT can be a powerful framework for effectively utilizing existing domain knowledge (heuristic solutions) and extracting superior performance in COP like TSP. This approach holds significant promise for developing new high-performance solution methods for real-world problems in logistics, manufacturing, and other fields where domain-specific solutions have been accumulated over many years.

\bibliographystyle{plain}
\bibliography{references}


\appendix \label{sec:appendix}

\section{Experimental details} \label{sec:appendix_experimental_details}

\subsection{Dataset generation details}

The TSP instance data were downloaded or generated using the code from the GitHub repository of \cite{learning-tsp} (\url{https://github.com/chaitjo/learning-paradigms-for-tsp}).
The breakdown is provided in Table \ref{tab:data-recipe}.

The TSP solutions by the NN, NI, and FI heuristics were also generated using the code from the GitHub repository of \cite{learning-tsp}.
The TSP solutions by SA were generated using the code from \url{https://github.com/perrygeo/simanneal}. The hyperparameters for SA are shown in Table \ref{tab:sa-recipe}.

\begin{table}[H]
    \centering
    \caption{Details of TSP dataset generation and sources.}
    \label{tab:data-recipe}
    \begin{tabular}{l|ccc}
        \toprule
        Data & $N=20$ & $N=50$ & $N=100$ \\
        \midrule
        Training & Download & Download & Download\\
        Validation & Download & Generate (Seed 9999) & Generate (Seed 9999)\\
        Test & Download & Download & Download\\
        \bottomrule
    \end{tabular}
\end{table}

\begin{table}[H]
    \centering
    \caption{Hyperparameters for SA used for generating solution data.}
    \label{tab:sa-recipe}
    \begin{tabular}{l|ccc}
        \toprule
        Parameter & $N=20$ & $N=50$ & $N=100$ \\
        \midrule
        Maximum (starting) temperature & 2.5 & 2.5 & 2.5\\
        Minimum (ending) temperature & 0.025 & 0.0025 & 0.0025\\
        Number of iterations & 50,000 & 5,000,000 & 5,000,000\\
        \bottomrule
    \end{tabular}
\end{table}

\subsection{Model architecture details} \label{sec:appendix_model_details}

The encoder was constructed based on the architecture of Kool et al. \cite{kool}. The detailed parameters are presented in Table \ref{tab:encoder-recipe}.

The decoder was built based on the implementation of the Elastic DT \cite{edt} (\url{https://github.com/kristery/Elastic-DT}). The detailed parameters are presented in Table \ref{tab:decoder-recipe}.

The parameters used for training are shown in Table \ref{tab:training-recipe}.

\begin{table}[H]
    \centering
    \caption{Architectural details of the transformer encoder.}
    \label{tab:encoder-recipe}
    \begin{tabular}{l c}
        \toprule
        Parameter & Value \\
        \midrule
        Number of layers & 2 \\
        Number of attention heads & 8 \\
        Embedding dimension & 128 \\
        Activation function & GELU \\
        Normalization method & Layer Normalization \\
        Dropout rate & 0.0 \\
        \bottomrule
    \end{tabular}
\end{table}

\begin{table}[H]
    \centering
    \caption{Architectural details of the Causal transformer decoder.}
    \label{tab:decoder-recipe}
    \begin{tabular}{l c}
        \toprule
        Parameter & Value \\
        \midrule
        Number of layers & 2 \\
        Number of attention heads & 8 \\
        Embedding dimension & 128 \\
        Activation function & GELU \\
        Normalization method & Layer Normalization \\
        Dropout rate & 0.0 \\
        Context length & Same as number of TSP nodes \\
        Reward clipping & False \\
        Expectile Regression quantile $\alpha$ & 0.99 \\
        \bottomrule
    \end{tabular}
\end{table}

\begin{table}[H]
    \centering
    \caption{Training hyperparameters.}
    \label{tab:training-recipe}
    \begin{tabular}{l c}
        \toprule
        Parameter & Value \\
        \midrule
        Loss balance coefficient $c$ & 0.5 \\
        Optimizer & Schedule-Free AdamW \cite{schedulefree} \\
        Learning rate & 0.0025 \\
        Weight decay & 0.0 \\
        AdamW betas & (0.9, 0.999) \\
        AdamW epsilon & 1e-8 \\
        Batch size & 1000 \\
        Maximum Epochs & 2000 \\
        \bottomrule
    \end{tabular}
\end{table}

\subsection{Computational environment} \label{sec:appendix_computation}

All experiments were conducted on a single server equipped with an NVIDIA GeForce RTX 4090 (24GB VRAM) and an Intel Xeon Silver 4314 CPU (2.40GHz). The models were implemented using PyTorch 2.5.1. The training and inference time for a single model is shown in the Table \ref{tab:computation}.

\begin{table}[H]
    \centering
    \caption{The computational time in training and predicting on NN dataset. The prediction time was measured as the duration required to predict a single instance.}
    \label{tab:computation}
    \begin{tabular}{l|ccc}
        \toprule
        Parameter & $N=20$ & $N=50$ & $N=100$ \\
        \midrule
        Training time & 19 hours & 1 days 3 hours & 1 days 23 hours\\
        Prediction time & 0.63 seconds & 0.86 seconds & 1.10 seconds\\
        \bottomrule
    \end{tabular}
\end{table}

\section{Detailed numerical results} \label{sec:appendix_numerical_results}

The actual average costs of the solutions for each method, used to calculate the optimality gaps in Tables \ref{tab:main}, \ref{tab:exp}, and \ref{tab:offset}, are presented below.

\begin{table}[H]
\centering
\caption{Actual average solution costs corresponding to the optimality gaps reported in Table \ref{tab:main}. The first row "Optimal" shows the average cost of the optimal solutions.}
\label{tab:act-main}
\begin{tabular}{ll|rrr}
\toprule
\multicolumn{1}{l}{Data} & \multicolumn{1}{l}{Method} & \multicolumn{1}{c}{$N=20$} & \multicolumn{1}{c}{$N=50$} & \multicolumn{1}{c}{$N=100$} \\
\midrule
\multirow{1}{*}{Optimal} & & 3.83 $\pm$ 0.30 & 5.69 $\pm$ 0.25 & 7.76 $\pm$ 0.23\\
\midrule
\multirow{3}{*}{NN} & Original & 4.49 $\pm$ 0.55 & 6.99 $\pm$ 0.57 & 9.69 $\pm$ 0.58\\
\cmidrule(lr){2-5}
 & BC & 4.49 $\pm$ 0.54 & 6.99 $\pm$ 0.57 & 9.69 $\pm$ 0.58\\
 & DT (Ours) & 4.47 $\pm$ 0.54 & 6.98 $\pm$ 0.56 & 9.69 $\pm$ 0.58\\
\midrule
\multirow{3}{*}{NI} & Original & 4.33 $\pm$ 0.39 & 6.78 $\pm$ 0.35 & 9.45 $\pm$ 0.33\\
\cmidrule(lr){2-5}
 & BC & 4.30 $\pm$ 0.39 & 6.73 $\pm$ 0.37 & 9.47 $\pm$ 0.44\\
 & DT (Ours) & 4.07 $\pm$ 0.33 & 6.54 $\pm$ 0.33 & 9.30 $\pm$ 0.40\\
\midrule
\multirow{3}{*}{FI} & Original & 3.92 $\pm$ 0.34 & 6.01 $\pm$ 0.32 & 8.36 $\pm$ 0.31\\
\cmidrule(lr){2-5}
 & BC & 3.90 $\pm$ 0.33 & 5.90 $\pm$ 0.30 & 8.17 $\pm$ 0.29\\
 & DT (Ours) & 3.88 $\pm$ 0.33 & 5.87 $\pm$ 0.29 & 8.13 $\pm$ 0.28\\
\midrule
\multirow{3}{*}{SA} & Original & 3.89 $\pm$ 0.33 & 5.94 $\pm$ 0.32 & 8.72 $\pm$ 0.36\\
\cmidrule(lr){2-5}
 & BC & 3.87 $\pm$ 0.32 & 5.86 $\pm$ 0.29 & 8.56 $\pm$ 0.43\\
 & DT (Ours) & 3.86 $\pm$ 0.32 & 5.83 $\pm$ 0.29 & 8.23 $\pm$ 0.31\\
\bottomrule
\end{tabular}
\end{table}

\begin{table}[H]
    \centering
    \caption{Actual average solution costs corresponding to the optimality gaps reported in Table \ref{tab:exp}.}
    \label{tab:act-exp}
    \begin{tabular}{ll|rr|rrr}
        \toprule
        & & \multicolumn{2}{c}{Fixed RTG} & \multicolumn{3}{c}{Predicted RTG using expectile regression} \\
        \multicolumn{1}{l}{$N$} & \multicolumn{1}{l}{Data} & \multicolumn{1}{c}{0} & \multicolumn{1}{c}{mean of data} & \multicolumn{1}{c}{$\alpha=0.50$} & \multicolumn{1}{c}{$\alpha=0.70$} & \multicolumn{1}{c}{$\alpha=0.99$} \\
        \midrule
        20 & NN & 6.43 $\pm$ 1.25 & 4.49 $\pm$ 0.54 & 4.49 $\pm$ 0.54 & 4.49 $\pm$ 0.54 & 4.47 $\pm$ 0.54 \\
         & NI & 6.44 $\pm$ 1.04 & 4.32 $\pm$ 0.23 & 4.33 $\pm$ 0.36 & 4.25 $\pm$ 0.36 & 4.07 $\pm$ 0.33 \\
         & FI & 7.25 $\pm$ 1.23 & 3.95 $\pm$ 0.26 & 3.90 $\pm$ 0.33 & 3.89 $\pm$ 0.32 & 3.88 $\pm$ 0.33 \\
         & SA & 7.88 $\pm$ 1.27 & 3.92 $\pm$ 0.26 & 3.87 $\pm$ 0.32 & 3.86 $\pm$ 0.32 & 3.86 $\pm$ 0.32 \\
        \midrule
        50 & NN & 17.84 $\pm$ 2.38 & 6.99 $\pm$ 0.56 & 6.98 $\pm$ 0.56 & 6.98 $\pm$ 0.57 & 6.98 $\pm$ 0.56 \\
         & NI & 14.78 $\pm$ 2.33 & 6.72 $\pm$ 0.32 & 6.71 $\pm$ 0.35 & 6.69 $\pm$ 0.35 & 6.54 $\pm$ 0.33 \\
         & FI & 11.55 $\pm$ 2.01 & 5.93 $\pm$ 0.26 & 5.91 $\pm$ 0.28 & 5.91 $\pm$ 0.28 & 5.87 $\pm$ 0.29 \\
         & SA & 20.41 $\pm$ 2.60 & 5.89 $\pm$ 0.24 & 5.87 $\pm$ 0.28 & 5.86 $\pm$ 0.28 & 5.83 $\pm$ 0.29 \\
        \bottomrule
    \end{tabular}
\end{table}

\begin{table}[H]
    \centering
    \caption{Actual average solution costs corresponding to the optimality gaps reported in Table \ref{tab:offset}. The applied offset values are shown in parentheses.}
    \label{tab:act-offset}
    \begin{tabular}{l|rrr}
        \toprule
        Data & \multicolumn{1}{c}{$N=20$ (offset)} & \multicolumn{1}{c}{$N=50$ (offset)} & \multicolumn{1}{c}{$N=100$ (offset)}\\
        \midrule
        NN & 4.44 $\pm$ 0.53 (1.00) & 6.97 $\pm$ 0.56 (1.00) & 9.67 $\pm$ 0.58 (2.00)\\
        NI & 3.98 $\pm$ 0.36 (0.50) & 6.28 $\pm$ 0.38 (2.00) & 9.04 $\pm$ 0.45 (2.00)\\
        FI & 3.88 $\pm$ 0.32 (-0.05) & 5.87 $\pm$ 0.29 (0.20) & 8.11 $\pm$ 0.28 (0.50)\\
        SA & 3.86 $\pm$ 0.32 (-0.10) & 5.83 $\pm$ 0.29 (0.00) & 8.18 $\pm$ 0.31 (0.50)\\
        \bottomrule
    \end{tabular}
\end{table}

\end{document}